\documentclass[conference]{IEEEtran}
\IEEEoverridecommandlockouts
\usepackage{cite}
\usepackage{amsmath,amssymb,amsfonts}
\usepackage{mathtools} 
\usepackage{subfig}
\usepackage{algorithmic}
\usepackage{graphicx}
\usepackage{hyperref}
\usepackage{algorithm2e}
\usepackage{textcomp}
\usepackage{xcolor}
\newcommand{\vecA}{\boldsymbol{\alpha}}
\newcommand{\bbE}{\mathbb{E}}
\newcommand{\vecX}{\mathbf{x}}

\def\BibTeX{{\rm B\kern-.05em{\sc i\kern-.025em b}\kern-.08em
    T\kern-.1667em\lower.7ex\hbox{E}\kern-.125emX}}
\begin{document}

\title{Beyond Simple Averaging: Improving NLP Ensemble Performance with Topological-Data-Analysis-Based Weighting
}

\author{\IEEEauthorblockN{1\textsuperscript{st} Polina Proskura}
\IEEEauthorblockA{\textit{Institute for Information Transmission Problems} \\
Moscow, Russia}
\and
\IEEEauthorblockN{2\textsuperscript{nd} Alexey Zaytsev}
\IEEEauthorblockA{\textit{Skolkovo Institute of Science and Technology} \\
\textit{Risk Management, Sber}\\
Moscow, Russia}
}


\maketitle

\begin{abstract}
In machine learning, ensembles are important tools for improving the model performance.
In natural language processing specifically, ensembles boost the performance of a method due to multiple large models available in open source. 
However, existing approaches mostly rely on simple averaging of predictions by ensembles with equal weights for each model, ignoring differences in the quality and conformity of models. 

We propose to estimate weights for ensembles of NLP models using not only knowledge of their individual performance but also their similarity to each other. By adopting distance measures based on Topological Data Analysis (TDA), we improve our ensemble.
The quality improves for both text classification accuracy and relevant uncertainty estimation.
\end{abstract}


\section{Introduction}
\label{sec:intro}

Models for Natural Language Processing (NLP) are numerous and vary in quality and efficiency. 
They are used for solving different important real-life problems, including text classification, translation to different languages, grammatical acceptance problems, etc. The examples are described in \cite{jrfm16070327} and \cite{LIU2019103318}. 

Ensembling is a well-proven way to improve the quality of machine learning (ML) models. 
It helps not only with classical ML approaches, like random forests in ~\cite{kozlovskaia2017deep}, but also with neural networks.
In practice, we observe that ensembling improves over a single model approach in terms of quality/efficiency tradeoff \cite{powerlaw}.
Particular implementations leverage characteristics of the loss function surface. 
Models that correspond to different local optima tend to exhibit errors on distinct segments of the dataset. 
So, combining these models enhances overall performance, compensating for the weaknesses of single models if they are diverse~\cite{loss_surface}.

Ensembling can improve performance not only in cases with several strong models but also with the addition of a weak model. 
In this case, the quality of the strong model can be improved with a proper weak model's weight~\cite{izmailov2018averaging}.

Two main aspects need to be included in the decision process for identifying weights for each model: the quality of the individual model and the diversity of considered models. Most of the existing approaches focus on the first part and result in lower final quality, specifically for transformer-based models for NLP. 
However, the diversity of the models is also important  \cite{xiao2022uncertainty} because it has a direct effect on the models correcting each other in the ensemble. 
In this work, we include diversity in the weighting process and take both aspects into account. 

The potential influence of highly correlated models must be reduced to allow other models in the ensemble to improve overall quality and build an effective ensemble without training additional neural networks, that is shown, for example, in \cite{vazhentsev-etal-2022-uncertainty}.

For the reasons mentioned above, in this research, we calculate the divergence between models using several different approaches and use them to adjust the weights for the individual models in the ensemble to improve the quality.

The current research suggests using topological features of the attention layers of the transformer models to estimate the similarity between models in the ensemble. The features are both interpretable and representative, which allows, at the same time, checking the values and getting the necessary precision of the estimation. The final weighting in the ensemble is based on these calculations and allows the achievement of a higher quality of the final prediction. 


Experiments showed improvement in quality and uncertainty compared to the models weighted without considering divergence. The baseline increase on the ensemble with equal weights varies from $0.3\%$ to $0.6\%$ compared to the best single model for different datasets. Our method with optimal weighting allowed for an additional improvement from $0.3\%$ to $0.7\%$ for different datasets compared to the ensemble with equal weights. Also, weighting using topological analysis of attention layer features shows lower uncertainty in the final model compared to other ensembles.

\section{Related work}

\paragraph{Ensembles}

Ensembles are a very popular and effective method in classical ML and also in neural network models. It increases the quality of the models, decreases uncertainty, and helps to reduce overfitting.

Classical ML uses ensembles to reduce the bias and/or variance of the models. For example, the random forest approach uses bagging~\cite{breiman1996bagging} or gradient boosting~\cite{schwenk2000boosting}, each new model in the ensemble corrects the errors of the previously added model. 

For neural networks, the optimization using stochastic gradient descent will end in different local minima of the loss function for different initializations as shown in \cite{loss_surface} and in \cite{DBLP:journals/corr/abs-2104-02395}. The total quality of the ensemble increases because different local minima of respective models tend to make errors on different parts of the sample as highlighted in \cite{MOHAMMED2023757}. Some works, like \cite{powerlaw} and \cite{chirkova2020deep}, explore the idea of the ensemble with a more efficient architecture of models being able to provide better quality than a single, more complex model with the same training time. 
Training models for ensemble requires more computer resources and time~\cite{izmailov2018averaging}.

\paragraph{Ensembles in NLP}
Ensembles applied to NLP problems are proven to be effective techniques to improve the quality of the final model, for example, in \cite{aniol2019ensemble}. One of the main reasons for that is the large amount of data required for each NLP problem~\cite{LIU2019103318, jrfm16070327, tacla00483}, and the process requires an effective combination of the data for final inference.

\paragraph{Attention layer features}
The most popular NLP models are transformers, complicated models with millions of parameters. Such models are usually hard to interpret. The attention mechanism can be used to interpret the model as a mapping of similarities between words in the sequence.
Different features of the attention layers can be used for a clearer and more understandable description of the model, for example \cite{barannikov2022representation, Kushnareva_2021, Levy2020.08.01.231639}. 
They can be used to measure the similarity between two models in an ensemble. One such feature is the barcode \cite{cherniavskii-etal-2022-acceptability}. This is a topological feature calculated based on the attention graph, representing the attention weights of the sample. It characterizes persistent features of the graph and describes its stability. 

\paragraph{Uncertainty estimation}

It can be complicated to assess real-life quality for many ML methods due to incomplete and noisy data. To better estimate the model's performance and check the reliability of the predictions, \cite{angelopoulos2022gentle} and \cite{xiao2022uncertainty} suggested the uncertainty estimation (UE) measure. 
It responds to errors and helps with misclassification detection tasks. 
The measure is important for detecting the ensemble parts with weaker predicting ability. 
Uncertainty can be measured in a number of ways described in \cite{vazhentsev-etal-2022-uncertainty}.

The methods can be used specifically for NLP problems and provide computationally cheap and reliable UE as in \cite{ulmer-etal-2022-exploring} for transformer neural networks, which are widely used to work with texts, for example, in \cite{xiao2018quantifying}. 
These methods measure uncertainty through the interpretability of the model or selective predictions. For modern neural networks with the Softmax output or Bayesian neural networks, UE is a capability based on the interpretation of the output as probabilities. Other methods are Stochastic Gradient Langevin Dynamics for LSTM and Monte Carlo Dropout for Transformers. Most of the methods are oriented for the one model UE and are not suitable for ensembles.

\paragraph{Research gap}

One of the ways to make the ensembling effective is to change the weights of the models to increase total quality without extra models or additional training. Most of the existing works focus on including the quality of each model in weighting algorithms~ \cite{izmailov2018averaging, 10190625, shahhosseini2020optimizing}. The divergence between models is often overlooked in such studies, although due to the nature of the ensembling, a high similarity between models often means a smaller quality increase from an additional model. 

This work considers the divergence of models in the weighting process by using interpretable and stable topological attention layer features.

\section{Methods}

Let us consider an ensemble of models $\{y_i(\vecX)\}_{i = 1}^k$ of $k$ models.
Predictions are aggregated to an ensemble with weights $\alpha_i$ as $\hat{y}(\vecX) = \sum_{i = 1}^{k} \alpha_i y_i(\vecX)$.

The standard approach is to apply equal weights $\alpha_i = \frac{1}{k}$ for all $i$. In our approach by adjusting weights $\alpha_i$ we are able to consider models similarities and pay less attention to inferior ones. Moreover, we propose how to utilize topological features to improve diversity estimation.

\subsection{Weighting}

Let us consider the regression problem.
We have $k$ neural network models in an ensemble; the ensemble has the form 
$\hat{y}(\vecX) = y_{\vecA} (\vecX) = \sum_{i = 1}^k \alpha_i y_i(\vecX)$
for $\alpha_i \geq 0$, $\sum_{i = 1}^k \alpha_i = 1$, where $y_i$ is the $i$-th model prediction, and $\alpha_i$ is the weight of the $i$-th model in the ensemble. 

Let $y(\alpha)$ be the ensemble of the models with the models' weights $\alpha$. We are interested in minimizing the quadratic risk of $y(\alpha)$:
\begin{align}
\label{eq:main_problem}
&\mathbb{E}_{p(\vecX)} (y_{\alpha}(\vecX) - y_{\mathrm{true}}(\vecX))^2 \rightarrow \min_{\vecA} \\
\nonumber
& \text{s.t.}\, \sum_{i = 1}^k \alpha_i = 1,\; \alpha_i \geq 0,\; i = \overline{1, k},
\end{align}
where $y_{\mathrm{true}}$ is the true target value, and $p(\vecX)$ is the data distribution.

Assuming models differ in quality, for $i = \overline{1, k}$ we have:
\begin{align}
\label{eq:assumptions}
 &\bbE (y_{\mathrm{true}}(\vecX) - y_i(\vecX))^2 = a_i^2, \\
 \nonumber
 &\bbE y_i(\vecX) y_j(\vecX) = b_{ij}, \; c_i = b_{ii}.
\end{align}
Here $a_i$ represents the error of the $i$-th model, and $b_{ij}$ represents the correlation between $i$-th and $j$-th models. 
To estimate all the coefficients $\mathbf{a} = \{a_i\}$, 
$B = \{b_{ij}\}$, $\mathbf{c} = \{c_i\}$, 
we use the validation sample. Using these coefficients we optimize the quadratic risk $L(\vecA)$ of $y_{\vecA}$ in the equation ~\eqref{eq:main_problem} under the given constraints.
Then,
\begin{align}
\label{eq:solved}
  &L(\vecA) = \bbE (y_{\vecA}(\vecX) - y_{\mathrm{true}}(\vecX))^2 = \bbE (\sum_{i = 1}^k \alpha_i y_i(\vecX) - y_{\mathrm{true}}(\vecX))^2=\\
  \nonumber
  &= \bbE y_{\mathrm{true}}(\vecX)^2 - 2 \sum_i \alpha_i \bbE y_{\mathrm{true}}(\vecX) y_i(\vecX) + \sum \alpha_i^2 \bbE y_i(\vecX)^2 +\\
  \nonumber &+ 2 \sum \alpha_i \alpha_j \bbE y_i(\vecX) y_j(\vecX) = \sum \alpha_i (\bbE y_{\mathrm{true}}^2(\vecX) -\\\nonumber
  &-2 \bbE y_{\mathrm{true}}(\vecX) y_i(\vecX) + \bbE y_i(\vecX)^2) - \sum \alpha_i \bbE y_i(\vecX)^2 +\\\nonumber
 &+ \sum \alpha_i^2 \bbE y_i(\vecX)^2 + 2 \sum \alpha_i \alpha_j \bbE y_i(\vecX) y_j(\vecX) = \\\nonumber
  &= \sum \alpha_i \bbE (y_{\mathrm{true}}(\vecX) - y_i(\vecX))^2 - \sum \alpha_i \bbE y_i(\vecX)^2 + \\\nonumber &+ \sum \alpha_i^2 \bbE y_i(\vecX)^2 + 2 \sum \alpha_i \alpha_j \bbE y_i(\vecX) y_j(\vecX) = \\\nonumber
  &= \sum \alpha_i a_i - \sum \alpha_i c_i + \vecA^T B \vecA = \\\nonumber &= \vecA^T (\mathbf{a} - \mathbf{c}) + \vecA^T B \vecA.
\end{align}
This is a quadratic optimization problem described at~\cite{boyd2004convex} with linear equality and inequality constraints that can be easily solved. In case of classification problem, it can be easily transformed to regression by interpreting outputs as probabilities.

\subsubsection{Optimization Algorithms}
Quadratic risk optimization problem~\ref{eq:solved} can be solved in different ways, including Bayesian optimization explained in \cite{frazier2018tutorial} and 
quadratic programming methods. 
We focus on the latter to better balance the computation time and precision of the final weights with library \cite{cvxopt}. The approach can be also applied to optimizing other parameters of the model. 

\subsection{Similarity measure for models}
The measure of the similarity between models is a crucial part of the proposed method.
These similarity values $b_{ij}$ can be calculated using the properties of the models or using features obtained with the validation set. 

\paragraph{Models' output correlation}
The simplest way to calculate correlation is using the output of the models for the specific validation dataset. 
For a binary classification, let's define $\mathbf{f}$ and $\mathbf{s}$ as output vectors of the first $f(\vecX)$ and the second $s(\vecX)$ models, respectively.
Then, the element of the correlation matrix related to these two models can be estimated via the correlation of two vectors 
$b_{fs} = \mathrm{corr}(f, s) \approx \mathrm{corr}(\mathbf{f}, \mathbf{s})$. 

\cite{watson2022agree} suggests that the correlation poorly reflects correspondence between models because they are trained to get the best results. 
Namely, the closer models are to the ideal case, the more correlated the outputs are, and we obtain a biased estimation of the correlations in such cases.
Moreover, correlations only partially reflect connections between models~\cite{kornblith2019similarity}.

\subsubsection{Topology Features-Based Divergence}

To better measure the similarity between models, we propose the divergence-based topological data analysis as it better preserves complex interdependencies between models~\cite{barannikov2022representation}.

Specifically, we chose $RTD$-based topological features (Representation Topology Divergence) described in~\cite{barannikov2022representation} for similarity computations. 
They are calculated using the algorithm via graphs and barcodes defined below. 

\paragraph{Barcodes}
Let's take a graph $G$ with weighted edges and a threshold $\tau$. If the weight of an edge is smaller than $\tau$, the edge is "deleted" from the graph. 
Increasing the value of $\tau$ produces a set of graphs with filtered edges compared to the original graph. 
Each of these graphs has different features, like edges, cycles and other structural properties. 
Each feature has two values of $\tau$: the value when the feature first appears and the value when the feature disappears. 
The set of existence intervals for all features is a barcode. 
It can be used to describe the stability of the graph properties. 

\begin{figure}[ht]
  \centering
\includegraphics[width=1.0\linewidth]{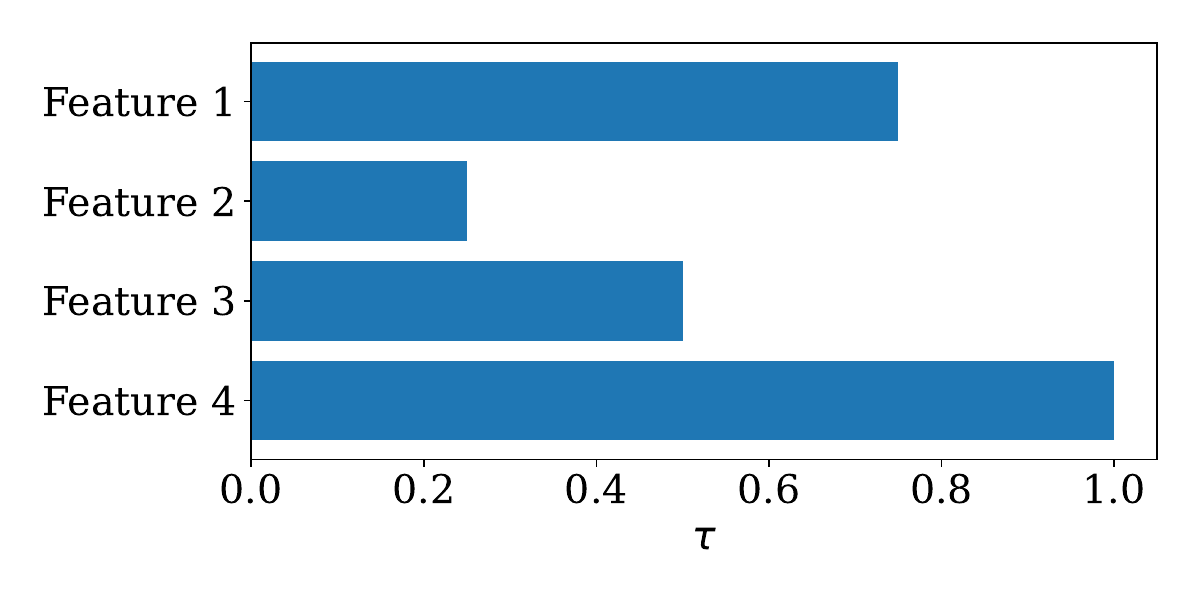}
  \caption{Example of Barcode for Four Features With Different Time $\tau$ of Deletion.}
    \label{fig:local_scheme3}
\end{figure}

The example of the barcode is in Figure~\ref{fig:local_scheme3}. 
The features $1-4$ are the existence of the edges of the graph $G$ depending on their weights. 
For \textit{Feature 1} 
the edge $1$ weights $0.75$, and for $\tau > 0.75$ the edge no longer exists in $G$. 

\begin{figure}[ht]
  \centering
\includegraphics[width=1.0\linewidth]{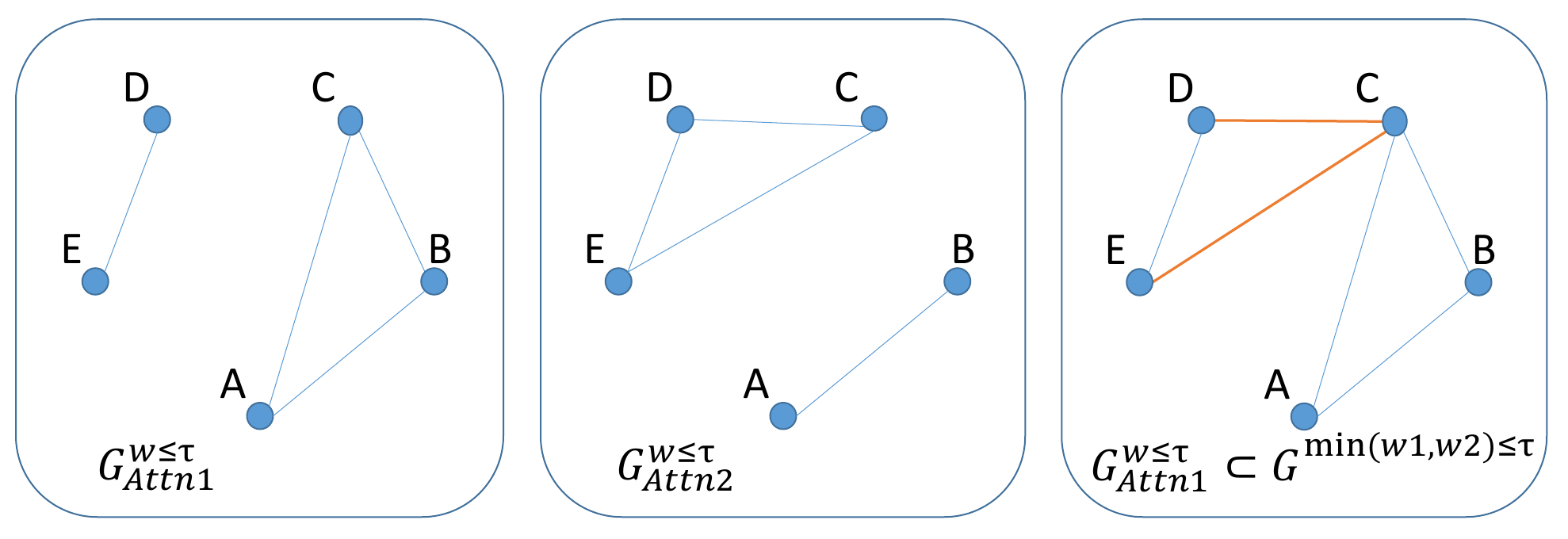}
  \caption{Example of Two Attention Graphs With Matched Vertices and Their Union Graph.}
    \label{fig:local_scheme4}
\end{figure}

\paragraph{R-Cross-Barcode} 
Let's define $A_{attn}$ as an attention layer matrix of a transformer model and graph $G$ as the weighted graph defined by $A_{attn}$ whose vertices are tokens and edges are the attention weights of the respective pairs of tokens. For two models, graphs $G_1$ and $G_2$ have the same vertex set. If we apply threshold $\tau$ with specific values for all edges, the graphs will transform to $G_1^{w_1 \leq \tau}$ and $G_2^{w_2 \leq \tau}$, where $w_1, w_2$ are edge weights in respective models. To calculate the similarity metric for $G_1^{w_1 \leq \tau}$ and $G_2^{w_2 \leq \tau}$, let's define $G_{union} = G^{\min (w_1, w_2) \leq \tau}$. In the example in Figure ~\ref{fig:local_scheme4}, the $G_{union}(\tau)$ is on the right and edges that are in $G_{union}(\tau)$ but not in $G_1^{w_1 \leq \tau}$ are orange. 

The measure of the similarity of $G_1^{w_1 \leq \tau}$ and $G_2^{w_2 \leq \tau}$ is calculated thought their individual similarity to $G_{union}(\tau)$. 
To measure the discrepancy between $G_1^{w_1 \leq \tau}$ and $G_{union}(\tau)$ we count the connected components of $G_1^{w_1 \leq \tau}$ that merged together in $G_{union}(\tau)$ by orange edges and its significance define by $\tau_1 - \tau_{union}$, where $\tau_1$ and $\tau_{union}$ are the smallest thresholds, that two components are merged in $G_1^{w_1 \leq \tau}$ and $G_{union}(\tau)$ respectively. The resulting R-Cross-Barcode is defined for all components by its specific $\tau_1$ and $\tau_{union}$. 

\paragraph{RTD} Finally, the RTD feature is calculated by averaging the sum of barcode intervals from several runs of R-Cross-Barcode calculation on randomized subsets of vertices for $G_1$ and $G_2$. 
The full algorithm for RTD calculation can be found in \cite{barannikov2022representation}. The RTD feature characterizes the degree of topological discrepancy between the representations and produces clearer divergence measures. In the research, it's calculated between the corresponding attention layers of each pair of ensemble models. 

\subsection{Algorithm}
The general description of the used algorithm is in the pseudocode~\ref{alg:main}.

\begin{algorithm}
\textbf{Input:} $M_1, \ldots ,M_k$ -- attention layer matrices for trained individual models. \\
$y_1, \ldots,y_k$ -- validation set outputs for respective models. \\
$\mathbf{a}$ -- error vector of the models, $a_i^2 = \bbE (y_{\mathrm{true}}(\vecX) - y_i(\vecX))^2$ for $i$-th model.\\
$R$ – the matrix of RTD divergence of each model pair, the values $R_{ij}$ will be calculated in the algorithm. \\
$i \gets 0, j \gets 0$\;
\For{$i\leq k$, $j \leq k$}
{
    $R_{i, j}\gets RTD(M_i, M_j)$\;
}
$\alpha_1, ... \alpha_k \gets Optimize(L(R, \mathbf{a}))$\;
where $L(R, \mathbf{a}) = \alpha^T(\mathbf{a} - \vec{R_{diag}}) + \alpha^TR\alpha$\\
\textbf{Result:} $\alpha_1, \ldots,\alpha_k$ -- normalized models weights in the ensemble.\\
$y_{final} = \sum_{i=1}^k \alpha_i \cdot y_i$  -- the final ensemble. \\
\caption{Weighting models in the ensemble using topology-based discrepancy}
\label{alg:main}
\end{algorithm}

\paragraph{General approach scheme}
\begin{itemize}
    \item[0.] Before calculating RTD features matrices are made symmetric with $M^{sym}=1-\max(M,M^T)$ as in Cherniavski \cite{cherniavskii-etal-2022-acceptability}.
    \item[1.] Calculate the similarity for each pair of the individual models to build a similarity matrix with correlation of the output vectors or using RTD features of the attention layer.
    \item[2.] Using the correlation matrix, find the optimal weights for the models in the ensemble using quadratic optimization.
    \item[3.] Use the weights for final ensemble prediction. 
\end{itemize}

A code is available at 
\href{https://anonymous.4open.science/r/NLP_ensembles-0B32/}{$NLP\_ensembles$}.
The code is written on PyTorch for easy integration and usage.
The repository shows how to reproduce the conducted experiments.

\section{Results}

\subsection{Experimental setup}

We consider our as well as alternative model weighting for a number of datasets.

\paragraph{Datasets}

The research uses four datasets, adopting the benchmark from~\cite{shelmanov-etal-2021-certain} for a complete analysis of the transformer's performance. 

\textbf{The IMDb dataset}~\cite{maas-etal-2011-learning} is a large collection of movie reviews used for binary sentiment classification, distinguishing between positive and negative reviews. It contains $50 000$ reviews split evenly into training and validation sets, with reviews preprocessed.
\textbf{The CoLA (Corpus of Linguistic Acceptability) dataset}~\cite{warstadt-etal-2019-neural} is designed for binary classification to determine the grammatical acceptability of English sentences. It comprises $10 657$ sentences labeled as grammatically acceptable or unacceptable. $9594$ sentences belong to the training sample, and the remaining $1063$ constitute the validation set.
\textbf{The SST-2 (Stanford Sentiment Treebank) dataset} ~\cite{socher-etal-2013-recursive} is a binary sentiment classification dataset that contains movie reviews. It consists of $~70000$ samples, with $~56\%$ positive and $~44\%$ negative reviews, the data is preprocessed.
Finally, \textbf{the MRPC (Microsoft Research Paraphrase Corpus) dataset}~\cite{dolan-brockett-2005-automatically} is a binary classification dataset for paraphrasing. Each sample contains two sentences and a label if they are paraphrases or not. There are $ 5801$ samples in the dataset, with $3900$ of them being paraphrases. 

\paragraph{Ensembles}
Bagging of neural networks is used for ensembling. $k$ single neural networks with the same architecture and different random initializations are trained. 
The prediction of an ensemble is $y = \sum_{i = 1}^k \alpha_i y_i$ for $\alpha_i \geq 0$, $\sum_{i = 1}^k \alpha_i = 1$, where $y_i$  and $\alpha_i$ are the predictions and the weight in the ensemble of the i-th model. In the research $k = 5$.
The weights of the ensemble can be found in different ways with two last approaches being proposed by us in this article:
\begin{itemize}
    \item \textit{Equal weights ensemble} has $\alpha_i = \frac{1}{k}$ for any $i$.
    \item \textit{Output correlation ensemble} calculates $\alpha_i$ in optimization using validation set output based correlation.
    \item \textit{Topology features ensemble} calculates $\alpha_i$ in optimization using correlation based on RTD features of attention layers.
    \item{\textit{Topology features ensemble (a subset of models)}} is similar to Topology features ensemble, but includes the selection of the subset of the original models set, varying number of models between $2$ and $5$ and choosing the models with the biggest weights.
\end{itemize}

\subsection{Metrics}
There are two main aspects that the research is focused on: the quality of the final ensemble and the quality of its uncertainty estimation (UE).
The quality measure is accuracy.
The UE quality measure is the area under the rejection curve (AURC) widely used recently~\cite{ding2020revisiting,geifman2019biasreduced,galil2021disrupting}. 
It is calculated in the following way: 
\begin{itemize}
    \item On the validation sample, the classifier returns the probabilities of the labels for each example.
    \item Our UE is MaxProb: the maximum of the probabilities returned by a classifier. We vary the threshold $\tau$ from $0$ to $1$, and for each value, discard samples with an uncertainty higher than $\tau$.
    \item The accuracy $a(\tau)$ is a curve that depends on the threshold and is called a rejection curve.
    \item AURC is the area under the rejection curve. 
\end{itemize}
The area under the rejection curve describes how uncertainty corresponds to the errors of a classifier, as shown in~\cite{rejectionCurve}.
Bigger areas correspond to better rejection and, thus, more accurate UE.

\subsection{Models}

The individual models for the ensemble are DistilBERTs \cite{sanh2020distilbert}. This is a well-performing version of BERT for NLP, which is lighter and faster than the original. 
A single model has $\approx 30k$ vocabulary size, $6$ layers of the encoder with $12$ heads of each layer, and GELU activation \cite{hendrycks2023gaussian}.
These models were previously used in ensembling and UE papers~\cite{risch-krestel-2020-bagging,shelmanov2021certain}. 
For additional weak model experiments, the \textit{lighter} version of the DistilBERT model with a single encoder layer is used.
If not specified, we use DistilBERT model.

Another base model is XLNet ~\cite{yang2020xlnet}. 
The model was pre-trained using an autoregressive method to learn bidirectional contexts and was proven to be effective and well-performing. 
A single model has $\approx 30k$ vocabulary size, $24$ layers of the encoder with $16$ heads of each layer, and also GELU activation \cite{hendrycks2023gaussian}. The model was previously used for different tasks ~\cite{9317379} and compared to other models ~\cite{cortiz2021exploring}. 

\subsection{Main results}

The obtained metrics are shown in Table~\ref{table:results}.
The inclusion of correlations in the process of finding optimal weights improves the quality of the resulting ensemble. 
Our dissimilarity scores based on topological features further improve both performance and the UE. 

\begin{table}[htbp]
    \caption{Accuracy and UE Quality for Different Combinations of the Models and Ensembling Types for IMDB, CoLA, SST-2 and MRPC Datasets.}
    \label{table:results}
    \centering
        \begin{tabular}{lllllll}
        \hline
        Model & Base  & Dataset & Accuracy & AURC\\  
              & model &  &  & \\  
        \hline
        Best single model & &  & $0.931$ & $0.982$\\ 
        Equal weights ensemble & DistilBERT & IMDB & $0.934$ & $0.984$\\
        Output correlation ensemble &  &  & $0.937$ & $0.978$\\
        Topology features ensemble &  &  & \textbf{0.937} & \textbf{0.986}\\
        \hline
        Best single model & & & $0.813$ & $0.902$\\ 
        Equal weights ensemble & DistilBERT & CoLA & $0.818$ & $0.907$\\
        Output correlation ensemble &  & & $0.819$ & $0.904$\\
        Topology features ensemble & & & \textbf{0.823} & \textbf{0.910} \\
        \hline
        Best single model &  & & $0.905$ & $0.973$\\ 
        Equal weights ensemble & DistilBERT & SST-2& $0.908$ & $0.975$\\
        Output correlation ensemble &  &  & $0.911$ & $0.974$\\
        Topology features ensemble &  &  & \textbf{0.913} & \textbf{0.976} \\
        \hline
        Best single model &  &  & $0.940$ & $0.987$\\ 
        Equal weights ensemble & XLNet & SST-2& $0.943$ & $0.988$\\
        Output correlation ensemble &  &  & $0.944$ & $0.987$\\
        Topology features ensemble &  & & \textbf{0.944} & \textbf{0.988} \\
        \hline
        Best single model &  &  & $0.838$ & $0.928$\\ 
        Equal weights ensemble & DistilBERT & MRPC& $0.843$ & $0.936$\\
        Output correlation ensemble &  & & $0.848$ & $0.935$\\
        Topology features ensemble &  &  & \textbf{0.850} & \textbf{0.941} \\
        \hline
        Best single model &  &  & $0.876$ & $0.960$\\ 
        Equal weights ensemble & XLNet & MRPC& $0.882$ & $0.966$\\
        Output correlation ensemble &  &  & $0.884$ & $0.966$\\
        Topology features ensemble &  &  & \textbf{0.885} & \textbf{0.969} \\
        \hline
        \end{tabular}
\end{table}

\paragraph{Rejection curves}

\begin{figure*}[ht]
  \centering
    \subfloat{{\includegraphics[width=0.47\linewidth]{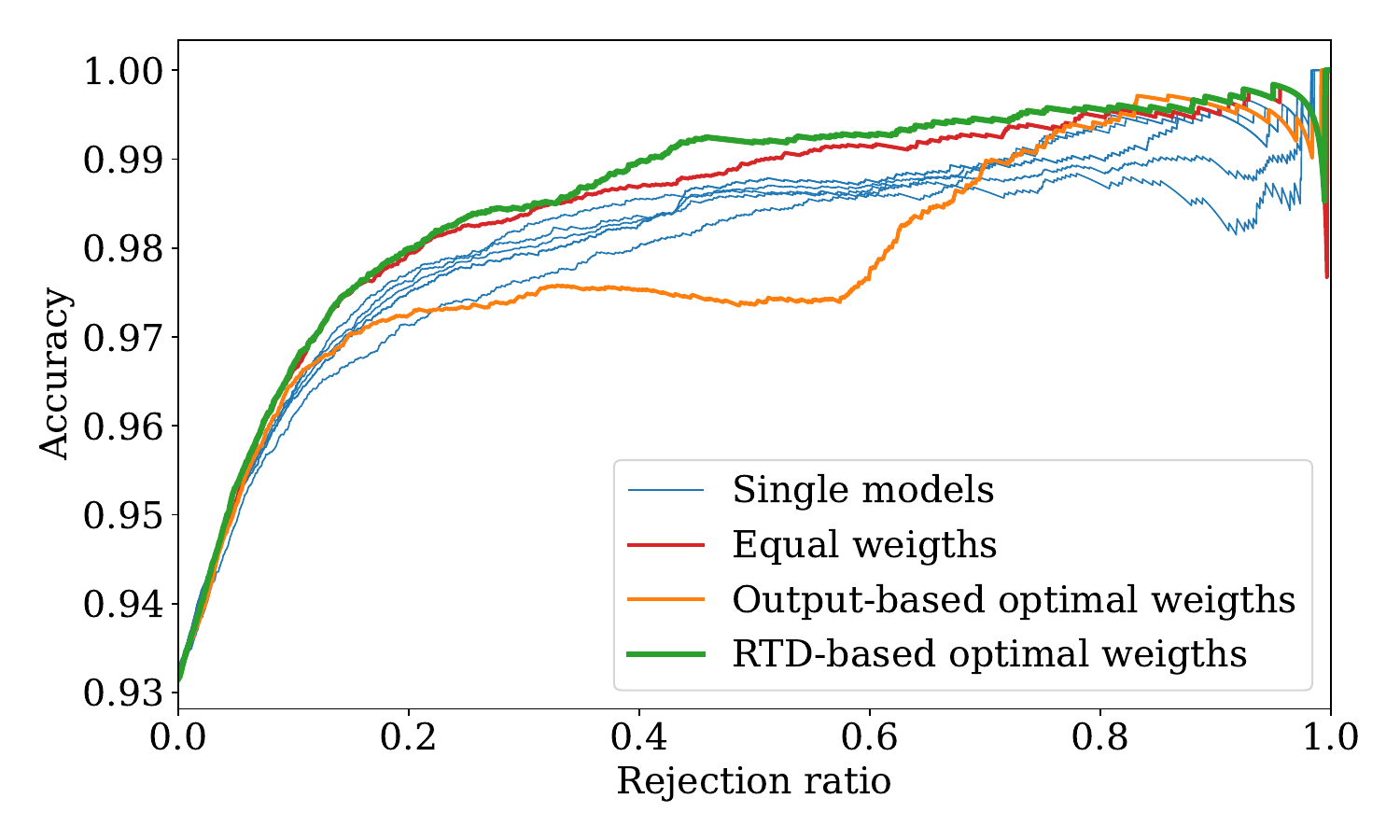} }}
    \qquad
    \subfloat{{\includegraphics[width=0.47\linewidth]{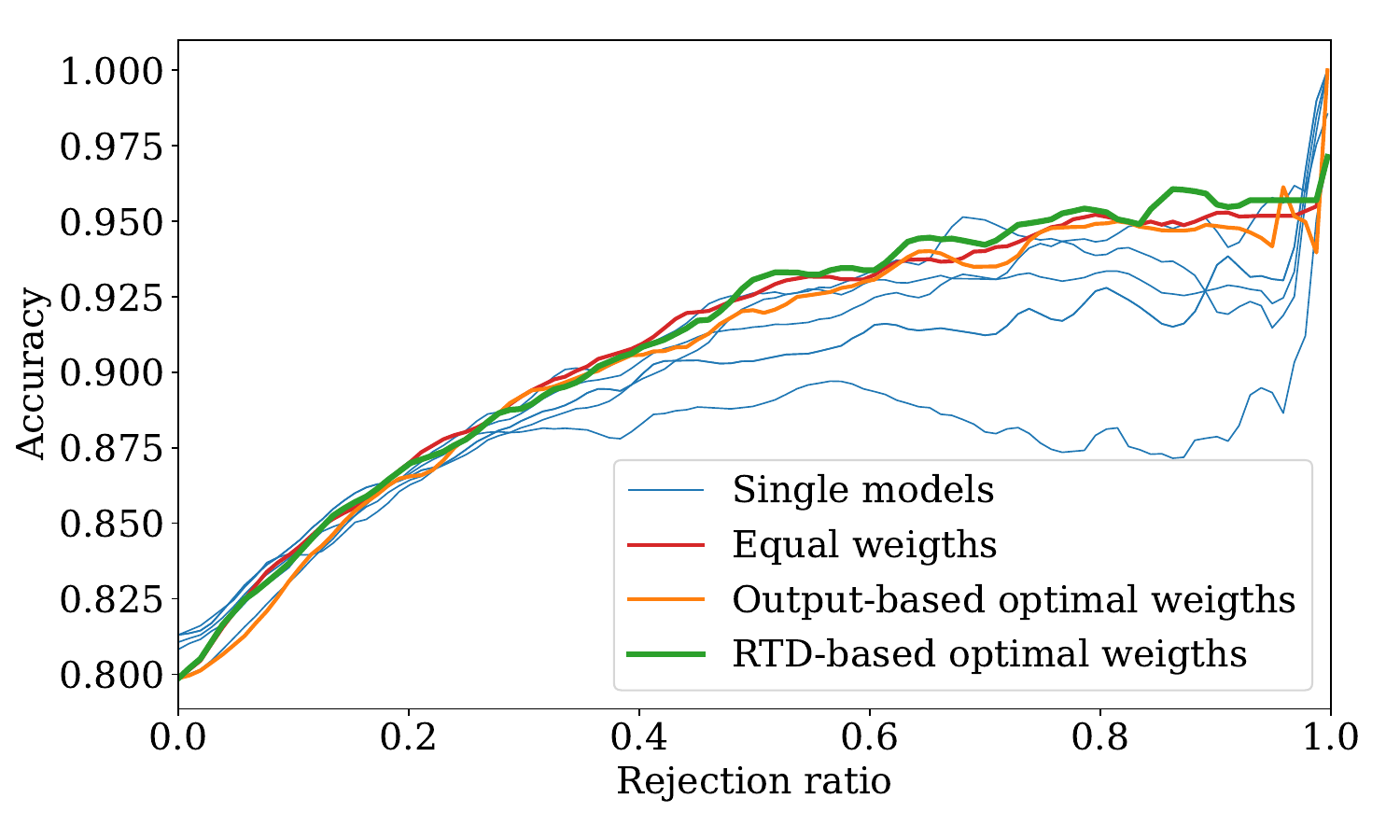} }}%
  \caption{Rejection Curves for IMDB (Left) and CoLA (Right) Datasets for DistilBERT model. The Blue curves correspond to Individual Models, the Red curve Is for Equal Weights Ensemble, the Orange Is for the Output-Based-Correlation Model and the Green Curve Is for the Ensemble Optimized Using the RTD-Based Correlation.}
    \label{fig:local_scheme1}
\end{figure*}

Figure~\ref{fig:local_scheme1} provides the rejection curves for the considered models.
The plots support the stated conclusion and show better quality and lower uncertainty for the ensemble that includes the correlation between models.


\subsection{Additional experiments}

One of the big parts of the research is pairs and subsets of models. 
It tests the hypothesis that a weaker model improves the quality of the ensemble and explores the use of a smaller number of models per ensemble for faster inference.

\paragraph{Subset of models}

The results in Table~\ref{table:results2} show the quality and uncertainty for selecting the best subset of the models (sizes from $2$ to $5$) from the original ensemble. 
Each subset was optimized individually. 

\begin{table}[htbp]
    \caption{Accuracy and UE Quality for Different Ensemble Subset Sizes for IMDB and CoLA Datasets.}
    \label{table:results2}
    \centering
        \begin{tabular}{lllll}
        \hline
        Number & \multicolumn{2}{c}{IMDB} & \multicolumn{2}{c}{CoLA} \\
        of models & Accuracy & AURC &Accuracy &AURC \\ 
        
        \hline
        5 & \textbf{0.937} & \textbf{0.986} &\textbf{0.823} & \textbf{0.910}\\
        4 & $0.937$ & $0.985$ & $0.820$ & $0.908$\\
        3 & $0.936$ & $0.983$ & $0.818$ & $0.907$\\
        2 & $0.935$ & $0.983$ & $0.816$ & $0.904$\\
        \hline
        \end{tabular}
\end{table}

We see that an independently optimized subset of models improves the quality compared to a single model. Still, the ensemble's uncertainty with the smaller number of models is much higher.

\paragraph{Weak \& strong pair of models}

The results in Table~\ref{table:results3} show that the combination of strong and weak models can improve the quality compared to the strong model.
The biggest improvement comes from our method called Topology features ensembles.
Thus, the method can identify and properly weight such a pair of models.

\begin{table}[htbp]
    \caption{Accuracy and UE Quality for Different Combinations of the Weak and Strong Models and Ensembling Types for IMDB and CoLA Datasets.}
    \label{table:results3}
    \centering
     
    \begin{tabular}{p{2.2cm} p{1.2cm}p{0.7cm}p{1.2cm}p{0.7cm}}
        \hline
        & \multicolumn{2}{c}{IMDB} & \multicolumn{2}{c}{CoLA} \\
        Model & Accuracy & AURC &Accuracy & AURC \\ 
        
        \hline
        Best strong model & $0.931$ & $0.982$ & $0.813$ & $0.902$\\
        Best weak model & $0.877$ & $0.894$ & $0.660$ & $0.750$\\
        Equal weights ensemble & $0.933$ & $0.981$ & $0.814$ & $0.893$\\
        Topology features ensemble & \textbf{0.935} & \textbf{0.983} & \textbf{0.815} & \textbf{0.904}\\
        \hline
        \end{tabular}
\end{table}

\subsection{Computational resources}
All the computations were performed using T4 GPU. Each ensemble contains $5$ models. The computation time of each part of the training process is measured, consisting of training single model $n$ times for ensembling and measuring the RTD optimization and weighting step. The additional time for the inference due to the usage of topological weighting is negligible.

\begin{table}[htbp]
    \caption{Computation time (minutes:seconds) for each step of the training process for Different Combinations of the Models for IMDB, CoLA, SST-2 and MRPC Datasets.}
    \label{table:comp}
        \centering
        \begin{tabular}{lllll}
        \hline
        Base model & Dataset & Ensemble with equal & Ensemble with optimal \\ 
         &  & weights, training time & weights, training time \\ 
        \hline
        DistilBERT & IMDB & 474:10 & 476:11 \\
        DistilBERT & CoLA & 92:45 & 94:37 \\
        DistilBERT & SST-2 & 172:25 & 174:37 \\
        XLNet & SST-2 & 246:40 & 248:18 \\
        DistilBERT & MRPC & 72:50 & 75:22 \\
        XLNet & MRPC & 59:10 & 61:02 \\
        \hline
        \end{tabular}
\end{table}

Table~\ref{table:comp} shows that time spent on weights optimization is small compared to training a $1$ extra single model for each researched case. That means that using the approach described in the paper more computationally efficient solutions can be achieved for improving neural network ensembles performance.

\subsection{Future work}
Future work on the topic will address the computational complexity of getting the ensemble itself by improving total training time via adjusting gradient descent or using snapshot ensembles. 

\section{Conclusion}
We proposed the procedure to optimize weights for the bagging of neural networks to increase the quality and the uncertainty estimation quality of an ensemble. 
The weights are obtained via computing topological similarities between models.

Our method of weighting increases the quality of the ensemble compared to an equal-weight ensemble. Weighting specifically with topological features-based correlation and optimizing for quality and uncertainty using quadratic risk, we achieved the best quality: accuracy increase compared to the best single model is $0.6\%$ for IMDB, $1.0\%$ for CoLA, $0.8\%$ for SST-2 and $0.9\%$ for MRPC datasets; accuracy increase compared to equal weights ensemble is $0.3\%$ for IMDB dataset, $0.5\%$ for CoLA, $0.5\%$ for SST-2 and $0.3\%$ for MRPC datasets. Also, we obtain the highest quality uncertainty estimate: $0.986$ AURC compared to $0.984$ for equal weights case and $0.978$ compared to using the output-based correlation.

Another finding of the research is that the introduced optimization method can be used to select the subset of neural networks in the ensemble with the best quality and uncertainty values. The algorithm is effective not only for a set of strong models but also for a combination of weak and strong models.

The optimization algorithms explored in the research can be used for more effective ensembling of neural networks. It’s possible to achieve a higher quality of the ensemble without allocating extra time and computing resources for training another model. Another use of the algorithms is to choose the most effective subset of the models in the ensemble for better quality and faster inference. This can potentially boost the research in the area by getting better results simpler and quicker. 

Current research can be expanded to ensembling of the models trained jointly,  where correlation between models is a much bigger problem and can be applied to machine translation and other problems. This way, specifically the training of neural networks can be improved in terms of computational resources required. 

Still, our methods demonstrated in this work are powerful instruments for effective neural network ensembling and they could improve many models performance.


\section*{Acknowledgment}
    The research was supported by the Russian Science Foundation grant 20-7110135.



\bibliographystyle{ieeetr}
\bibliography{references_short}


\end{document}